\documentclass[10pt,twocolumn,letterpaper]{article}

\usepackage{cvpr}              
\usepackage[accsupp]{axessibility}  

\usepackage{array}
\usepackage[table]{xcolor}
\usepackage{booktabs}
\usepackage{multirow}

\definecolor{cvprblue}{rgb}{0.21,0.49,0.74}
\usepackage[pagebackref,breaklinks,colorlinks,allcolors=cvprblue]{hyperref}

\def\paperID{5466} 
\def\confName{CVPR}
\def\confYear{2026}

\title{Multi-Modal Image Fusion via Intervention-Stable Feature Learning }

\author{
  Xue Wang\textsuperscript{1,2}, \quad Zheng Guan\textsuperscript{1*},\quad Wenhua Qian\textsuperscript{1*}, \quad Chengchao Wang\textsuperscript{1},\quad Runzhuo Ma\textsuperscript{3}\\
 {\normalsize \textsuperscript{1} School of Information Science and Engineering, Yunnan University}\\
  {\normalsize \textsuperscript{2} School of Artificial Intelligence, Nanyang Normal University}\\
    {\normalsize \textsuperscript{3} Department of Electrical and Electronic Engineering, Hong Kong Polytechnic University}\\
  {\tt\small  gz\_627@sina.com \quad whqian@ynu.edu.cn}
}

\begin{document}
\maketitle
\begin{abstract}

Multi-modal image fusion integrates complementary information from different modalities into a unified representation. Current methods predominantly optimize statistical correlations between modalities, often capturing dataset-induced spurious associations that degrade under distribution shifts. In this paper, we propose an intervention-based framework inspired by causal principles to identify robust cross-modal dependencies. Drawing insights from Pearl's causal hierarchy, we design three principled intervention strategies to probe different aspects of modal relationships: i) complementary masking with spatially disjoint perturbations tests whether modalities can genuinely compensate for each other's missing information, ii) random masking of identical regions identifies feature subsets that remain informative under partial observability, and iii) modality dropout evaluates the irreplaceable contribution of each modality. Based on these interventions, we introduce a Causal Feature Integrator (CFI) that learns to identify and prioritize intervention-stable features maintaining importance across different perturbation patterns through adaptive invariance gating, thereby capturing robust modal dependencies rather than spurious correlations. Extensive experiments demonstrate that our method achieves SOTA performance on both public benchmarks and downstream high-level vision tasks. \textit{ The  \href{https://github.com/wang-x-1997/Multi-Modal-Image-Fusion-via-Intervention-Stable-Feature-Learning/tree/main}{Code} can be available.}

\end{abstract}

\section{Introduction}
Multi-modal image fusion (MMIF) aims to integrate complementary information from different sensing modalities into a unified representation that is more informative and reliable than any individual modality alone \cite{1,2,3,4,15,16}. In infrared and visible image fusion (IVIF), a subtask of MMIF, texture-rich structural detail from the visible spectrum is fused with semantic and thermal cues from infrared sensing. The fused output typically exhibits higher perceptual quality and richer scene, and is more resilient under degradation conditions, thereby supporting downstream high-level vision tasks \cite{7,8,55,56}. As a result, MMIF has become a key component in applications such as security monitoring, autonomous driving, and medical imaging \cite{5,6,9,14}.

\begin{figure}[t!]
    \centering
    \includegraphics[scale=0.32]{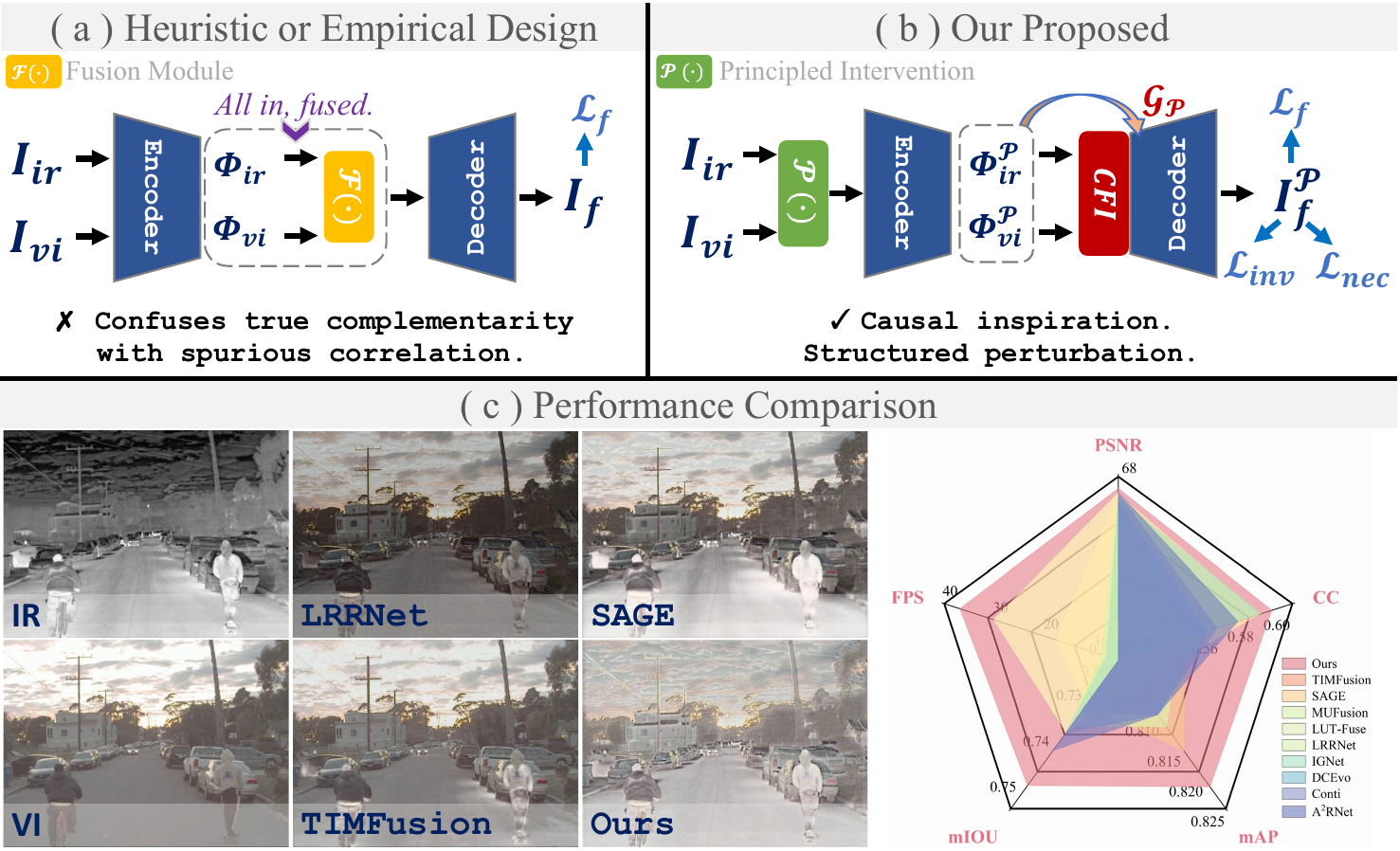}
    \caption{
Comparison with SOTA method in training framework and performance.
(a) General designs of existing methods, which often rely on empirical trial-and-error to fit all source features.
(b) Our framework actively probes modal dependencies through structured interventions.
(c) The superiority of our method is validated by its advantages in static metrics (CC, PSNR), efficiency analysis (FPS), semantic segmentation (mIoU), object detection (mAP), and qualitative comparisons.
    }
    \label{Fig1}
\end{figure}

The evolution of deep learning has catalyzed remarkable progress in MMIF. Current SOTA methods leverage sophisticated architectures, from CNN-based dual-stream networks to Transformer-based global attention mechanisms, to model complex inter-modal relationships \cite{10,11,12,13}. These approaches typically formulate fusion as an optimization problem that maximizes statistical dependencies between modalities, employing various losses to preserve intensity, gradient, and semantic information from source images. Recent advances further incorporate task-specific guidance, adversarial training, and diffusion models to enhance fusion quality \cite{3,4,6,46,37}.

Despite these empirical successes, current MMIF methods share a fundamental limitation: they learn from observational data without distinguishing genuine complementary relationships from spurious statistical regularities. This correlation-centric paradigm creates several critical challenges. When thermal signatures consistently co-occur with specific visible patterns in training data, models capture these statistical associations regardless of whether they reflect meaningful dependencies or dataset artifacts. Consequently, features are selected based on co-occurrence frequency rather than their actual contribution to fusion quality. This superficial learning leads to brittle models that fail when deployment conditions differ from training distributions, as the spurious correlations they rely upon no longer hold.
The root of these issues lies in the passive nature of observational learning. Models trained solely on input-output pairs cannot determine whether observed correlations are causal or coincidental. This echoes a fundamental distinction in machine learning: while correlation reveals patterns in data, understanding causation requires active intervention. Pearl's causal hierarchy \cite{17} formalizes this insight through three levels of reasoning: association (\emph{observing patterns}), intervention (\emph{manipulating variables}), and counterfactual (\emph{imagining alternatives}). \textbf{Current MMIF methods operate exclusively at the association level, missing crucial insights that intervention-based reasoning could provide.} However, applying causal principles to MMIF is nontrivial: unlike standard supervised tasks, fusion lacks explicit supervision for feature preservation, and modality complementarity often violates common independence assumptions. These properties demand a careful design of causal reasoning to the fusion setting.

Motivated by this gap, we propose a novel framework that leverages causal insights to design principled interventions for robust multi-modal fusion. We draw inspiration from causal principles to systematically probe modal dependencies through structured perturbations. \textbf{\emph{Our core idea is that features genuinely important for fusion should maintain their relevance across different intervention patterns, while spurious correlations should break down under systematic perturbation.}}
We instantiate this idea through three complementary intervention strategies, each designed to test specific hypotheses about modal relationships. \textbf{Complementary masking} applies spatially disjoint perturbations to different modalities, testing whether one modality can genuinely compensate for missing information in the other. This reveals true cross-modal complementarity as opposed to redundant encoding. \textbf{Random masking} corrupts identical regions across modalities to identify feature combinations that remain informative despite partial observability, highlighting robust local dependencies. \textbf{Modality dropout} completely removes individual modalities to quantify their irreplaceable contributions, preventing over-reliance on any single information source.

Building on these interventions, we introduce the Causal Feature Integrator (CFI), which identifies intervention-invariant features through adaptive invariance gating. Unlike conventional attention mechanisms that weight features based on statistical salience, CFI explicitly models regions that exhibit consistent importance across diverse perturbation patterns. This mechanism enables the network to prioritize robust cross-modal dependencies while suppressing spurious correlations that arise from dataset biases. Our key contributions are summarized as follows:
\begin{itemize}
\item We propose a systematic intervention framework for MMIF that moves beyond passive correlation learning to actively probe modal dependencies, inspired by principles from causal reasoning to identify robust fusion patterns.

\item We design three principled intervention strategies that test complementary aspects of modal relationships: cross-modal compensation, local sufficiency, and global necessity, each addressing specific failure modes in correlation-based fusion.

\item We introduce the CFI with learnable invariance gating, which explicitly identifies and aggregates features that remain stable across interventions, enabling more robust and interpretable fusion decisions.

\item Through extensive experiments on multiple benchmarks and downstream tasks, we demonstrate that intervention-based training produces models with superior generalization, particularly under challenging conditions where correlation-based methods fail.
\end{itemize}

\section{Related Work}
\subsection{Learning-based MMIF}
Deep learning has driven substantial progress in multi-modal feature processing through adaptive feature extraction, fusion, and reconstruction \cite{1,18,19,20,21,38}. Classical autoencoder-based methods rely on trained encoders and decoders to ensure effective feature representation, while enforcing cross-modal complementarity via carefully designed fusion rules \cite{22,23,39,40}. Representative methods such as CDDFuse \cite{2} integrate Transformer and CNN architectures to jointly capture global structures and local details. Similarly, SwinFusion \cite{10} adopts an end-to-end architecture that combines Transformer and CNN modules, enabling adaptive feature fusion and reducing reliance on manually engineered heuristics.
Beyond these architectures, generative fusion methods such as TarDAL \cite{9} formulate fusion as an adversarial game between the fused image and source modalities, introducing task-level supervision and optimizing fusion under multi-task objectives. More recently, diffusion models have further improved fusion quality \cite{41,42}. For example, Mask-DiFuser \cite{4} leverages generative diffusion to produce high-quality multi-modal fused images, while ControlFusion \cite{5} combines text conditioning with diffusion to achieve controllable, degradation-aware fusion.

\subsection{Causal-Inspired Learning}
The distinction between correlation and causation has motivated numerous works to incorporate causal principles into machine learning, improving generalization, robustness, and interpretability. Causal reasoning has achieved notable success in tasks such as low-light enhancement \cite{24,25}, self-supervised representation learning \cite{26}, and domain generalization \cite{27,28}. These methods typically employ intervention-based strategies, counterfactual augmentation, or structural causal models to identify stable relationships that generalize beyond training distributions.
In multi-modal learning, causal perspectives have been explored to achieve robust feature learning by modeling inter-modal relationships \cite{50,54} and employing intervention techniques to mitigate spurious correlations \cite{51,52,53}. 



\section{Method}
This section presents our intervention-based fusion framework, using IVIF as the primary instantiation with natural extensions to other multi-modal fusion scenarios.

\subsection{Problem Formulation}
Let $I_{vi} \in \mathbb{R}^{3 \times H \times W}$ and $I_{ir} \in \mathbb{R}^{1 \times H \times W}$ denote the visible and infrared input images, with $I_{f} \in \mathbb{R}^{1 \times H \times W}$ representing the fused output. Conventional fusion methods model this process through statistical optimization:
\begin{equation}
    I_{f} = \mathcal{F}(I_{ir}, I_{vi}) + \mathbf{n}_f,
\end{equation}
where $\mathcal{F}(\cdot,\cdot)$ represents the learned fusion mapping and $\mathbf{n}_f$ captures model uncertainty. These approaches minimize reconstruction losses to capture statistical dependencies, treating all correlations as potentially informative. However, this passive learning paradigm cannot distinguish genuine modal complementarity from spurious co-occurrences arising from dataset biases.

To address this limitation, we propose a different perspective inspired by causal reasoning principles. Consider the underlying data generation process: a latent scene $S$ generates observations through different modalities $S \rightarrow \{\Theta_{ir}, \Theta_{vi}\} \rightarrow I_{f}$, where $\Theta_{ir}$ and $\Theta_{vi}$ represent modal-specific features. External factors such as lighting conditions and sensor characteristics create spurious correlations that do not reflect true modal dependencies. To identify robust fusion patterns, we need to move beyond passive observation to active probing.
Drawing inspiration from Pearl's causal hierarchy \cite{17}, we recognize three levels of understanding that could benefit fusion: observing correlations (\emph{association}), testing under perturbations (\emph{intervention}), and reasoning about alternatives (\emph{counterfactual}).  Our key insight is that features genuinely important for fusion should maintain their relevance under principled perturbations, while spurious correlations should degrade. This motivates our intervention-based approach: systematically perturbing inputs through structured masking operations $\mathcal{M}$ to identify intervention-stable features.

\subsection{Principled Intervention Design}
We design three complementary intervention strategies, each testing specific hypotheses about modal relationships and addressing different failure modes of correlation-based fusion.

\textbf{Intervention 1: Testing Cross-Modal Compensation.} 
True modal complementarity means modalities can compensate for each other's missing information, not just co-occur statistically. Based on this principle, we apply spatially disjoint perturbations through complementary masking. We generate non-overlapping masks $\mathcal{M}^v, \mathcal{M}^i \in \{0,1\}^{H \times W}$ satisfying $\mathcal{M}^v \cap \mathcal{M}^i = \textbf{O}$, where each mask randomly occludes multiple spatial regions. The intervention produces:
\begin{equation}
    I_{f}^{c} = \mathcal{F}(I_{ir} \odot \mathcal{M}^i, I_{vi} \odot \mathcal{M}^v),
\end{equation}
where $\odot$ denotes element-wise multiplication. If the model successfully reconstructs the scene despite disjoint corruptions, it demonstrates genuine cross-modal compensation rather than within-modal memorization. \emph{This intervention specifically probes whether information from modality $\mathcal{A}$ can functionally substitute for missing content in modality $\mathcal{B}$.}

\textbf{Intervention 2: Identifying Locally Sufficient Features.} 
Robust fusion should tolerate partial observability, maintaining quality even when local regions are corrupted. To identify such locally sufficient feature combinations, we apply identical random masks to both modalities:
\begin{equation}
    I_{f}^{r} = \mathcal{F}(I_{ir} \odot \mathcal{M}^r, I_{vi} \odot \mathcal{M}^r),
\end{equation}
where $\mathcal{M}^r \in \{0,1\}^{H \times W}$ randomly occludes identical spatial locations across modalities. Features that preserve fusion quality despite these perturbations represent robust local dependencies essential for scene understanding. \emph{This intervention reveals which spatial regions contain sufficient information for high-quality fusion, independent of their statistical frequency in training data.}

\textbf{Intervention 3: Quantifying Modal Necessity.} 
Over-reliance on single modalities is a common failure mode when models exploit spurious correlations. To ensure balanced modal utilization, we perform complete modality dropout:
\begin{equation}
    I_{f}^{i} = \mathcal{F}(I_{ir} \odot \textbf{O}, I_{vi}), \quad 
    I_{f}^{v} = \mathcal{F}(I_{ir}, I_{vi} \odot \textbf{O}).
\end{equation}
This extreme intervention quantifies each modality's irreplaceable contribution by measuring performance degradation under complete removal. \emph{By enforcing substantial quality loss when either modality is absent, we prevent the model from learning shortcuts that ignore complementary information.}

These three interventions work synergistically: complementary masking ensures genuine cross-modal interaction, random masking identifies robust local patterns, and modality dropout prevents degenerative solutions. Together, they guide the model toward learning intervention-stable features that reflect true modal dependencies.

\begin{figure}[t!]
    \centering
    \includegraphics[scale=0.31]{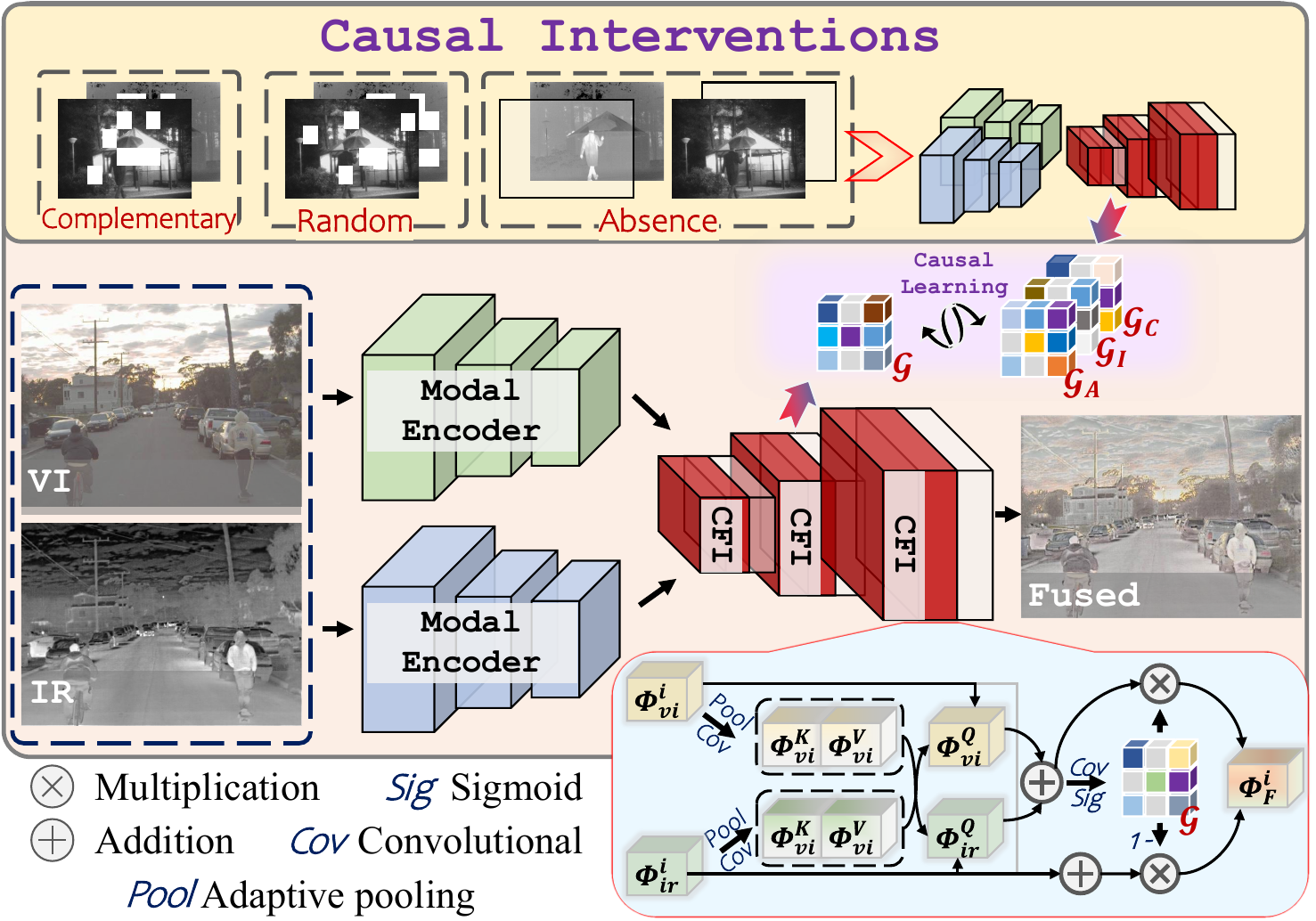}
    \caption{
The framework of the proposed method. It employs a U-Net-like Siamese architecture that incorporates CFI within the decoder to enable robust multi-modal fusion. By leveraging three complementary intervention strategies, the model learns to identify intervention-stable features that represent genuine cross-modal complementarity rather than spurious statistical co-occurrences arising from dataset biases.
    }
    \label{Net}
\end{figure}

\subsection{Network Architecture}

Our architecture, illustrated in Figure \ref{Net}, implements the intervention framework through a U-Net-based design with specialized fusion modules. The network consists of parallel encoders for feature extraction and a decoder with our proposed Causal Feature Integrator (CFI) for intervention-aware fusion.

\textbf{Feature Extraction.} Two weight-sharing encoders process $I_{vi}$ and $I_{ir}$ independently, each containing three convolutional blocks with progressive downsampling. This generates multi-scale representations $\{\Theta^v_1, \Theta^v_2, \Theta^v_3\}$ and $\{\Theta^i_1, \Theta^i_2, \Theta^i_3\}$ that capture both fine details and semantic abstractions. The parallel design preserves modality-specific characteristics while enabling subsequent cross-modal reasoning.

\textbf{Causal Feature Integrator (CFI).} 
The CFI module learns to identify and prioritize intervention-stable features through adaptive fusion. For features $\{\Theta_k^i, \Theta_k^v\}$ at scale $k$, CFI performs three operations: cross-modal attention for information exchange, invariance gating for stability assessment, and adaptive fusion for feature integration.

First, we compute bidirectional cross-modal attention to capture complementary relationships. We generate \emph{queries}, \emph{keys}, and \emph{values} through $1\times1$ convolutions:
\begin{equation}
\begin{aligned}
Q^v_k, K^v_k, V^v_k &= \mathrm{Conv}_{1\times1}(\Theta^v_k), \\
Q^i_k, K^i_k, V^i_k &= \mathrm{Conv}_{1\times1}(\Theta^i_k).
\end{aligned}
\end{equation}

To manage computational cost, we pool keys and values to compact representations of size $r \times r$ where $r \ll H, W$.  This pooling strategy preserves three critical properties: \emph{i}) spatial continuity without hard boundaries, \emph{ii}) global receptive fields essential for handling spatial misalignment between modalities, and \emph{iii}) efficient scaling to high-resolution multi-scale features. Cross-modal attention then computes:
\begin{equation}
\begin{aligned}
\Theta^{v \rightarrow i}_k &= \mathrm{Attention}(Q^v_k, \mathrm{Pool}(K^i_k), \mathrm{Pool}(V^i_k)), \\
\Theta^{i \rightarrow v}_k &= \mathrm{Attention}(Q^i_k, \mathrm{Pool}(K^v_k), \mathrm{Pool}(V^v_k)),
\end{aligned}
\end{equation}
where $\Theta^{v \rightarrow i}$ captures infrared information relevant to visible queries and vice versa. This bidirectional mechanism enables the network to reason about what each modality uniquely contributes.

\begin{figure}[t!]
    \centering
    \includegraphics[scale=0.32]{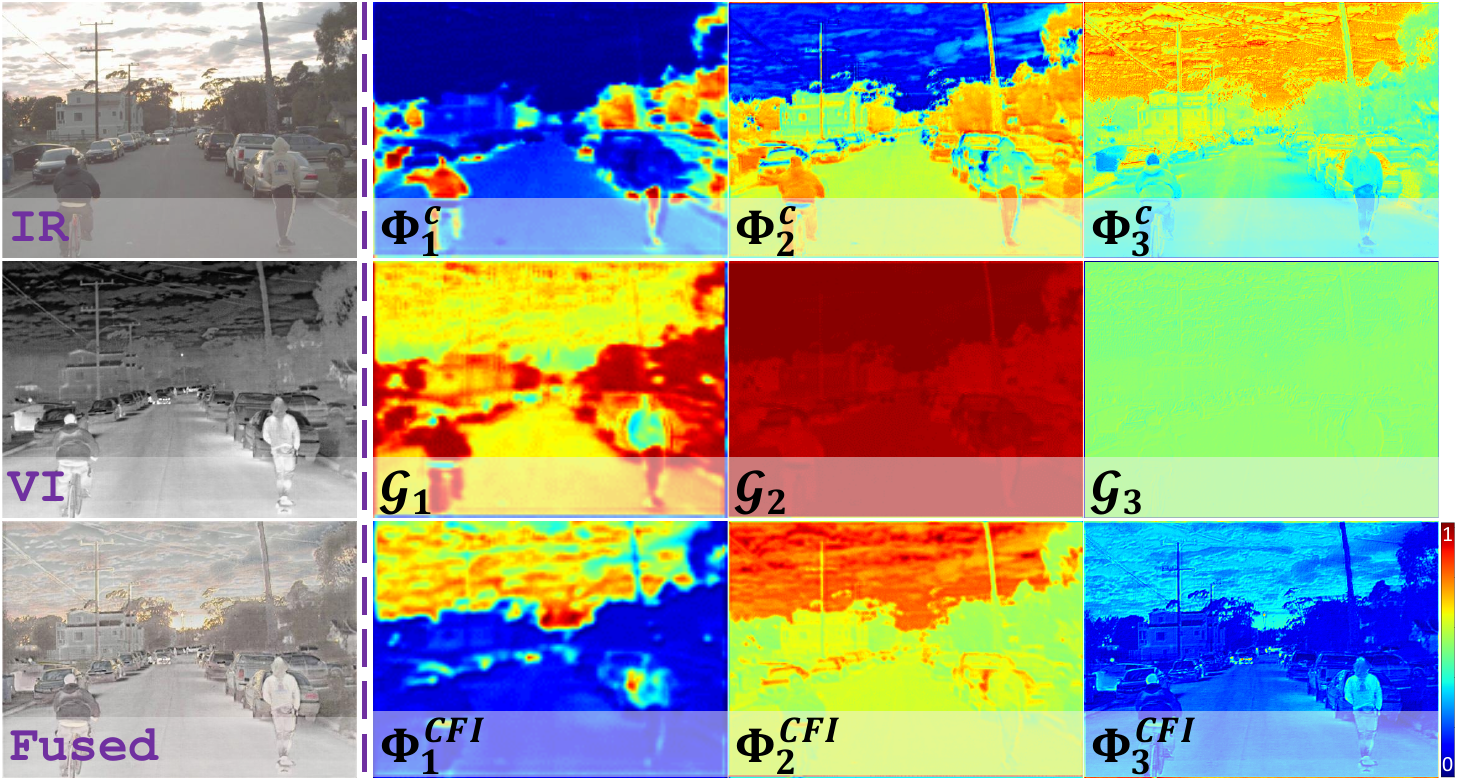}
    \caption{
An illustrative example of feature visualization. Through the learnable invariance gates, the model progressively focuses on intervention-stable features across successive iterations, prioritizing regions that maintain consistent importance under perturbations to ensure effective integration of complementary modalities.
    }
    \label{valization}
\end{figure}

Next, we aggregate complementary and local features:
\begin{equation}
\Theta^{c}_k = \Theta^{v \rightarrow i}_k + \Theta^{i \rightarrow v}_k, \quad
\Theta^{l}_k = \Theta_k^i + \Theta_k^v,
\end{equation}
where $\Theta^{c}_k$ represents cross-modal complementary features and $\Theta^{l}_k$ contains local modal information.

To emphasize intervention-stable information, a learnable invariance gate $\mathcal{G}_k$ is employed to blend the two feature types:
\begin{equation}
\Theta_k^{\mathrm{CFI}} = \mathcal{G}_k \odot \Theta^{c}_k + (1 - \mathcal{G}_k) \odot \Theta^{l}_k,
\end{equation}
where $\mathcal{G}_k = \sigma(\mathrm{Conv}_{3\times3}(\Theta_k^{c}))$ with sigmoid activation $\sigma(\cdot)$. High gate values indicate regions that remain informative across interventions (stable features), while low values mark intervention-sensitive areas (potentially spurious). This gating mechanism explicitly models feature stability, enabling the network to prioritize robust dependencies over spurious correlations.

\textbf{Feature Reconstruction.} The decoder progressively refines features through upsampling and skip connections, integrating CFI outputs at each scale. This hierarchical design enables both local detail preservation and global semantic consistency, with intervention stability propagating across scales. Figure \ref{valization} visualizes how CFI progressively focuses on genuine ntervention-stable features across scales, achieving interpretable fusion decisions based on intervention stability rather than statistical saliency.

\begin{figure*}[t!]
    \centering
    \includegraphics[scale=0.5]{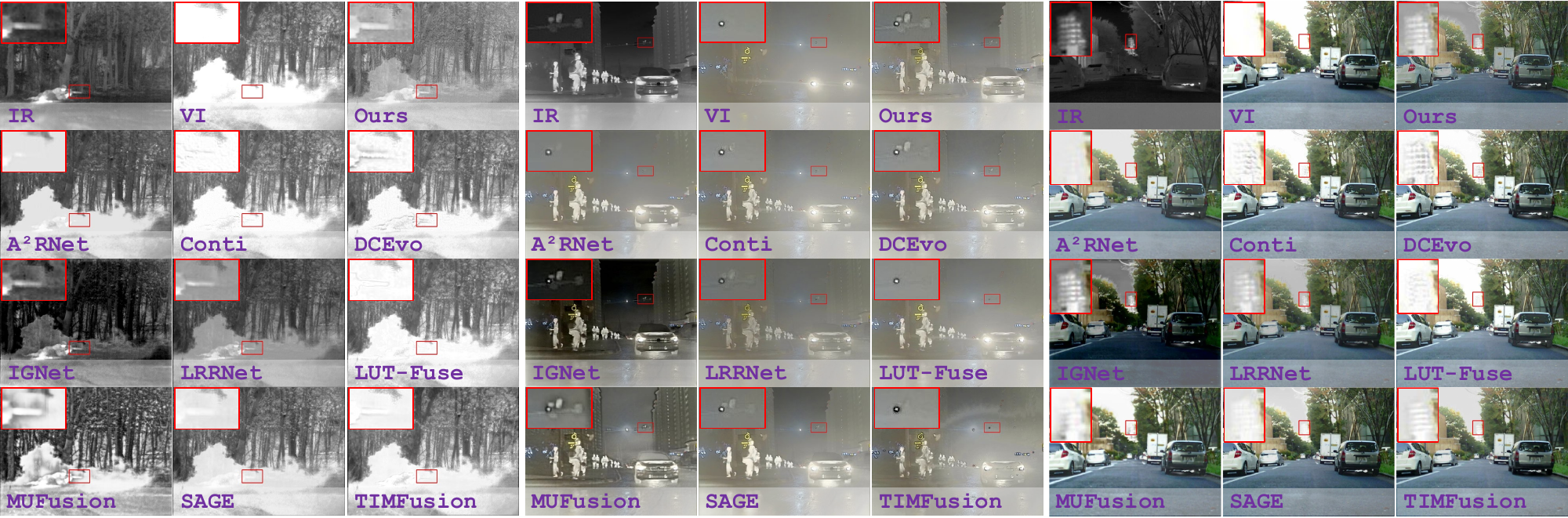}
    \caption{
Qualitative comparison with SOTA methods on the IVIF benchmarks.
    }
    \label{Dataset}
\end{figure*}

\begin{table*}[t!]
\centering
\caption{ Quantitative results of our proposed method \emph{vs.} SOTA methods on the IVIF benchmarks. The best value is highlighted with {\textbf{Bold}}.}
\label{tab:comparison}
\small
\setlength{\tabcolsep}{4pt}
\begin{tabular}{l*{15}{c}}
\toprule
\multirow{2}{*}{\textbf{Method}} & \multicolumn{5}{c}{\textbf{TNO}} & \multicolumn{5}{c}{\textbf{MSRS}} & \multicolumn{5}{c}{\textbf{M$^3$FD}} \\
\cmidrule(lr){2-6} \cmidrule(lr){7-11} \cmidrule(lr){12-16}
& \textbf{AG} & \textbf{SF} & \textbf{PSNR} & \textbf{CC} & \textbf{$\mathcal{Q}_{abf}$} 
& \textbf{AG} & \textbf{SF} & \textbf{PSNR} & \textbf{CC} & \textbf{$\mathcal{Q}_{abf}$} 
& \textbf{AG} & \textbf{SF} & \textbf{PSNR} & \textbf{CC} & \textbf{$\mathcal{Q}_{abf}$} \\
\midrule
TIMFusion    & 4.019 & 3.901 & 61.41 & 0.379 & 0.397 & 3.747 & 4.378 & 59.48 & 0.525 & 0.424 & 4.272 & 5.176 & 61.85 & 0.425 & 0.489 \\
Conti        & 3.860 & 3.987 & 61.12 & 0.446 & 0.437 & 3.737 & 4.504 & 64.26 & 0.603 & 0.570 & 4.476 & 5.545 & 61.11 & 0.479 & 0.499 \\
SAGE         & 3.817 & 3.745 & 59.81 & 0.317 & 0.444 & 3.431 & 4.373 & 65.59 & 0.594 & 0.529 & 4.590 & 5.443 & 61.88 & 0.557 & 0.510 \\
MUFusion     & 4.713 & 3.867 & 61.73 & 0.448 & 0.319 & 3.160 & 3.482 & 63.85 & 0.589 & 0.422 & 4.220 & 4.155 & 61.72 & 0.451 & 0.372 \\
LUT-Fuse     & 3.810 & 3.830 & 61.23 & 0.436 & 0.403 & 3.801 & 4.494 & 64.24 & 0.602 & \textbf{0.579} & 4.080 & 5.019 & 61.31 & 0.470 & 0.446 \\
LRRNet       & 3.855 & 3.806 & 61.72 & 0.436 & 0.346 & 2.672 & 3.309 & 64.68 & 0.515 & 0.447 & 3.613 & 4.212 & \textbf{62.95} & 0.541 & 0.484 \\
IGNet        & 4.198 & 3.857 & 60.48 & 0.459 & 0.368 & 3.318 & 3.889 & 65.33 & \textbf{0.655} & 0.455 & 4.716 & 5.137 & 60.55 & 0.525 & 0.400 \\
DCEvo        & 3.942 & 4.015 & 61.24 & 0.460 & \textbf{0.447}
& 3.807 & 4.512 & 64.49 & 0.605 & 0.620  &4.575 &5.517 &61.33 &0.500 &0.504  \\ 
A$^2$RNet       & 3.292 & 3.429 & 61.38 & 0.462 & 0.289 & 2.924 & 3.493 & 63.37 & 0.603 & 0.416 & 3.005 & 3.470 & 62.12 & 0.519 & 0.302 \\
\midrule
\textbf{Ours} & \textbf{5.128} & \textbf{5.132} & \textbf{62.06} & \textbf{0.502} & 0.423 & \textbf{4.129} & \textbf{4.743} & \textbf{66.02} & 0.646 & 0.545 & \textbf{5.276} & \textbf{6.105} & 62.13 & \textbf{0.565} & \textbf{0.512} \\
\bottomrule
\end{tabular}
\end{table*}

\subsection{Training Objective}
During the training phase, the model simultaneously performs principled intervention, outputting four intervention results ($I_f^{c}$, $I_f^{r}$, $I_f^{i}$, $I_f^{v}$), which are used to impose objective loss constraints. Our training objective combines three components to optimize fusion quality and learn intervention-stable patterns: \textbf{I.} Fusion fidelity loss $\mathcal{L}_{f}$ for perceptual fidelity to source modalities; \textbf{II.} Intervention consistency loss $\mathcal{L}_{\text{inv}}$ for feature stability under perturbations; \textbf{III.} Modal necessity loss $\mathcal{L}_{\text{nec}}$ to avoid over-dependence on single modalities:
\begin{equation}
\mathcal{L} = \mathcal{L}_{f} + \alpha\mathcal{L}_{\text{inv}} + \beta\mathcal{L}_{\text{nec}},
\end{equation}
where $\alpha$ and $\beta$ balance stability constraints against fusion quality.

\textbf{Fusion Fidelity Loss.} We use a composite objective to preserve intensity and structural details \cite{43,44}:
\begin{equation}
\begin{aligned}
\mathcal{L}_{f} = &\|I_f - I_{vi}\|_1 + \|I_f - I_{ir}\|_1 \\
&+ \lambda_1 \big\|\nabla I_f - \max(\nabla I_{vi}, \nabla I_{ir})\big\|_1,
\end{aligned}
\end{equation}
where $\nabla$ is the Laplacian operator and $\lambda_1$ is a trade-off coefficient. This ensures the fused image retains features from both modalities.

\textbf{Intervention Consistency Loss.} To enforce stability in key regions across perturbations, we apply:
\begin{equation}
\label{eq:inv_loss}
\mathcal{L}_{\text{inv}} = \sum_{I_j \in \{I_f^c, I_f^r\}} \!\! {\|(I_j - I_f)\odot\bar{\mathcal{G}}\|_1} + \mathcal{R}(\bar{\mathcal{G}}),
\end{equation}
where $I_f^c$ and $I_f^r$ are outputs under complementary and random masking, and $\bar{\mathcal{G}} = (\mathcal{G}_1 + \mathcal{G}_2 + \mathcal{G}_3)/3$ aggregates multi-scale gates. The first term penalizes differences in gate-selected stable regions.

The regularization $\mathcal{R}(\bar{\mathcal{G}})$ avoids degenerate solutions:
\begin{equation}
\mathcal{R}(\bar{\mathcal{G}}) = \|\mu(\bar{\mathcal{G}} - \eta)\|_1 - \mathbf{H}(\bar{\mathcal{G}}),
\end{equation}
where $\mu(\cdot)$ is spatial mean, $\eta$ targets moderate activation, and $\mathbf{H}(\cdot)$ is spatial entropy, promoting binary-like decisions for interpretability.
This loss trains gates to identify invariant features, focusing on robust cross-modal dependencies rather than spurious correlations.

\textbf{Modal Necessity Loss.} To ensure balanced modality use:
\begin{equation}
\label{eq:necessity}
\mathcal{L}_{\text{nec}} = \|I_f - I_f^i\|_1 + \|I_f - I_f^v\|_1,
\end{equation}
where $I_f^i$ and $I_f^v$ are infrared-only and visible-only outputs. This maximizes differences from single-modal fusions, preventing over-reliance on one modality, encouraging complementary feature discovery, and regularizing against biases.
The combined objective drives the model toward high-quality fusion with stable, complementary cross-modal patterns.

\section{ Experiment}
\subsection{Implementation Details}
Our experiments were conducted using PyTorch on a workstation equipped with an NVIDIA GeForce RTX 4090 GPU. We employed the training data provided by MSRS \cite{30} as the training dataset and RoadScene \cite{29} as the validation dataset. Training image pairs were cropped into $256\times256$ patches with random augmentation and fed into the network with a batch size of 16. The model was trained for 50 epochs using the Adam optimizer with a learning rate of 1e-4.  We empirically set the hyperparameters as $\alpha = 0.1$, $\beta = 0.05$, and $\lambda_1 = 1.0$  to ensure comparable magnitudes among the loss terms while prioritizing fusion quality as the primary objective. Following parameter analysis \emph{(reported in the Supplementary Materials due to space constraints)}, we set both the size of Complementary and Random masks to \(16\times16\); the number of masks was randomly sampled from 1 to 6 with the total masked area constrained not to exceed the training patch, and we set $\eta=0.3$ and $r=8$.
Five quantitative metrics, including PSNR, SF, AG, CC, and $\mathcal{Q}_{abf}$, were employed to objectively evaluate the fusion performance. Higher values indicate superior fusion results, with computational details provided in \cite{1}.

\subsection{Infrared and visible image fusion}
We evaluate our proposed method on three widely-used IVIF benchmarks: MSRS, TNO \cite{45}, and M$^3$FD \cite{9}. Our method was compared with \textbf{9} SOTA IVIF methods, including TIMFusion \cite{31}, Conti \cite{32}, SAGE \cite{33}, MUFusion \cite{34}, LUT-Fuse \cite{14}, LRRNet \cite{35}, IGNet \cite{36}, DCEvo \cite{3}, and A$^2$RNet \cite{13}.

\begin{figure}[t!]
    \centering
    \includegraphics[scale=0.23]{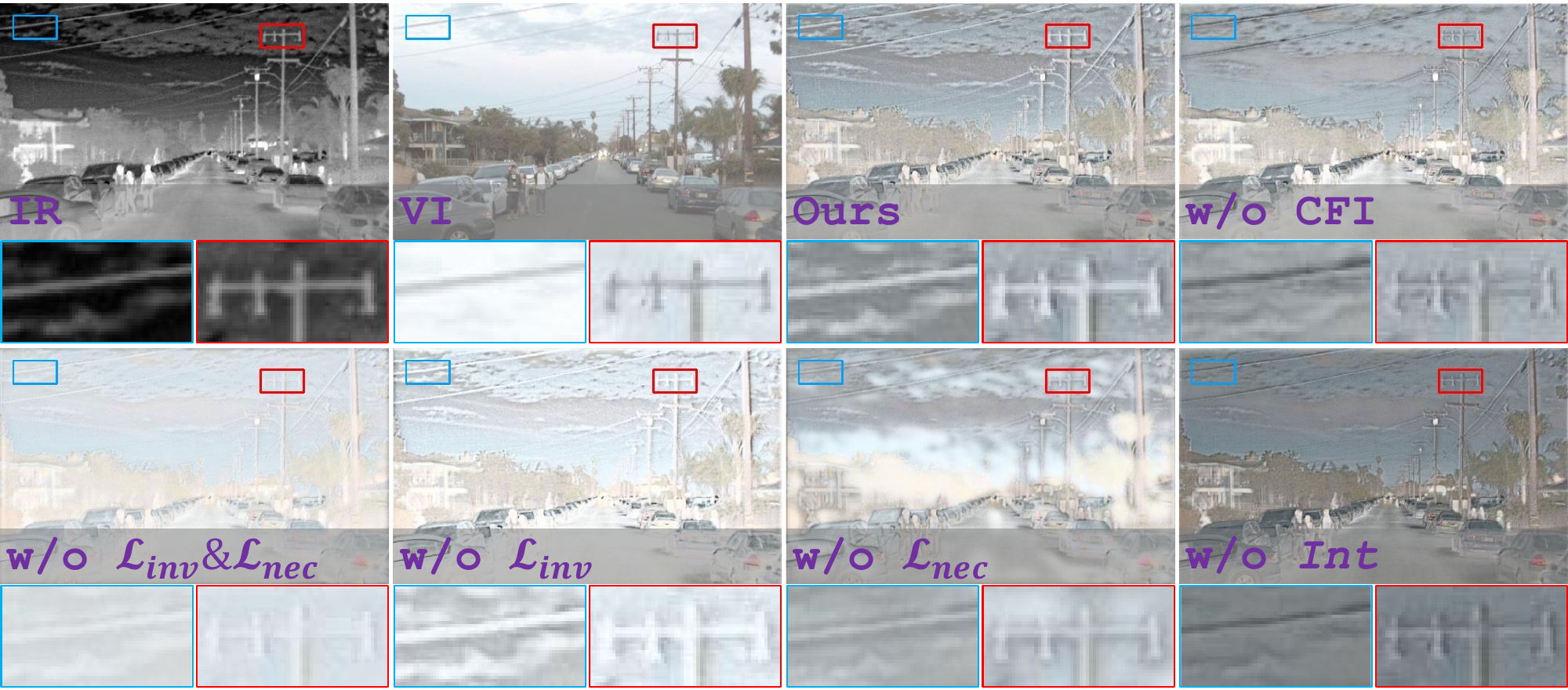}
    \caption{
Visualization comparison of ablation study
    }
    \label{Ablation}
\end{figure}

\subsubsection{Comparison with SOTA methods}
\textbf{Qualitative comparison.}
Figure \ref{Dataset} presents qualitative comparisons across three benchmarks. Through systematic intervention-based training, our method robustly integrates complementary features under extreme conditions such as heavy fog and overexposure, demonstrating significant advantages. In contrast, LRRNet's lightweight design and hand-crafted parameters limit its representation capacity, while semantically-guided methods like SAGE and DCEvo often produce blurred details or insufficient contrast in thermal regions due to their reliance on statistical correlations. By learning intervention-stable features through our principled perturbation strategies, our method achieves clear and accurate representation of thermal objects, effectively preserving critical semantic information.

\begin{table}[t!]
\centering
\caption{Ablation experiment results. The best value is highlighted with {\textbf{Bold}}.}
\label{tab:ablation}
\small
\setlength{\tabcolsep}{6pt}
\begin{tabular}{l*{5}{c}}
\toprule
\textbf{Case} & \textbf{AG} & \textbf{SF} & \textbf{PSNR} & \textbf{CC} & \textbf{$\mathcal{Q}_{abf}$} \\
\midrule
w/o CFI        & 5.764 & 5.972 & 60.21 & 0.544 & 0.428 \\
w/o $\mathcal{L}_{inv}$     & 5.179 & 5.728 & 58.08 & 0.573 & 0.331 \\
w/o $\mathcal{L}_{nec}$     & 4.016 & 4.018 & 61.39 & 0.393 & 0.368 \\
w/o $\mathcal{L}_{nec} \&\mathcal{L}_{inv}$       & 3.361 & 3.478 & 59.85 & 0.524 & 0.312 \\
w/o $Int$ & 5.332 & 5.348 & \textbf{63.95} & 0.598 & \textbf{0.524} \\
\midrule
\textbf{Ours} & \textbf{6.136} & \textbf{6.244} & 63.62 & \textbf{0.605} & 0.467 \\
\bottomrule
\end{tabular}
\end{table}

\textbf{Quantitative Comparisons.}
Table \ref{tab:comparison} validates the systematic advantages of our intervention framework, achieving optimal overall performance across three benchmarks. The intervention consistency constraints enhance edge preservation, cross-modal compensation through complementary masking ensures high-fidelity reconstruction in complex scenes, and the invariance gating mechanism achieves stable cross-scenario generalization. Compared to correlation-based methods, our intervention-guided approach delivers stronger robustness and generalization.

\subsubsection{Ablation Study} 
The ablation study in Table~\ref{tab:ablation} and Figure~\ref{Ablation} systematically validates each component's contribution. Removing CFI maintains edge metrics but introduces visible noise and structural distortions, confirming its role in identifying intervention-stable features. Without $\mathcal{L}_{inv}$, PSNR degrades substantially, demonstrating that intervention consistency constraints are essential for robust feature learning. Removing $\mathcal{L}_{nec}$ significantly impairs AG and SF with evident image distortion, validating its role in ensuring balanced modal utilization.
 Notably, completely eliminate intervention mechanisms and only use $\mathcal{L}_f$ and backbone network (w/o \emph{Int}) setting attains higher PSNR and $\mathcal{Q}_{abf}$ but markedly lower AG and SF, revealing a core trade-off in fusion goals. Correlation-driven optimization favors pixel fidelity and local gradients, whereas AG and SF capture structural coherence and texture richness crucial for semantics. As shown in Figure~\ref{Ablation}, w/o \emph{Int} yields smoother outputs but loses fine thermal details and edge definition. Our intervention framework instead prioritizes structural integrity and feature stability, achieving more balanced performance across metrics.

\subsubsection{Intervention Impact Analysis}
\begin{figure}[t!]
    \centering
    \includegraphics[scale=0.35]{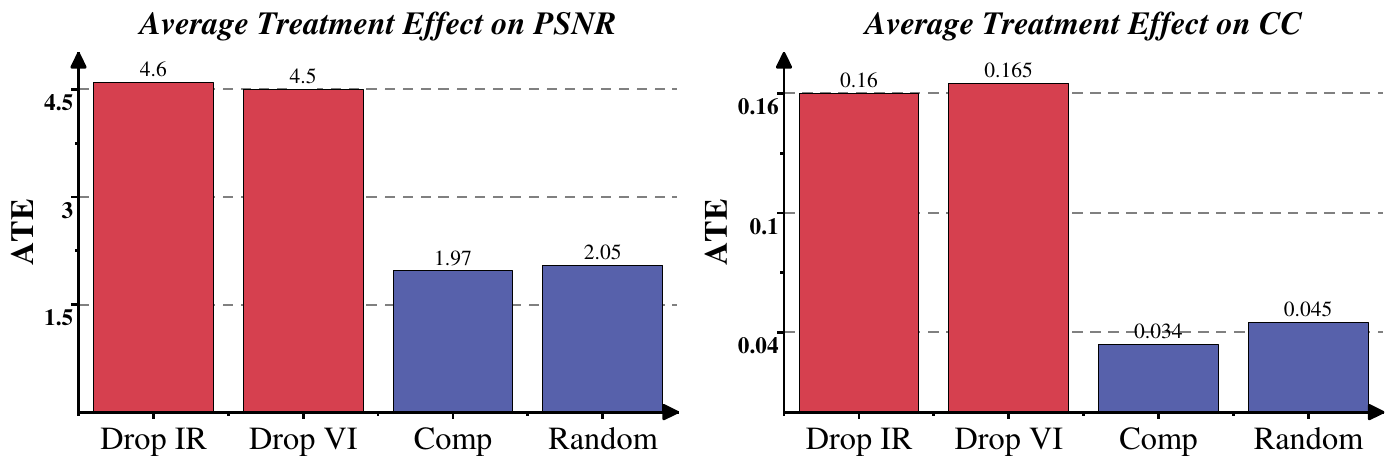}
    \caption{
Quantitative results of ATE.
    }
    \label{ATE}
\end{figure}

We quantify the impact of our interventions through Average Treatment Effect (ATE)~\cite{49} analysis. For intervention $T \in \{0,1,2,3,4\}$ (baseline, complementary/random masking, IR/VI dropout) and outcome $Y_i(t)$ representing fusion quality, ATE is defined as $\mathbb{E}[Y_i(0) - Y_i(t)]$. We estimate ATE via sample average: $\widehat{\text{ATE}}(t) = N^{-1}\sum_{i=1}^N [Q(f(I_i)) - Q(f(M_t(I_i)))]$, where $Q$ measures PSNR/CC and $M_t$ applies intervention $t$.
Figure~\ref{ATE} reveals distinct intervention impacts. Modality dropout induces the largest degradation, confirming both modalities provide irreplaceable information. Random masking produces minimal effects, indicating successful learning of locally sufficient features. Complementary masking shows moderate impact, validating cross-modal compensation capabilities. This confirms our framework learns intervention-stable features from modal-specific information to local patterns to cross-modal relationships.

\begin{figure}[t!]
    \centering
    \includegraphics[scale=0.22]{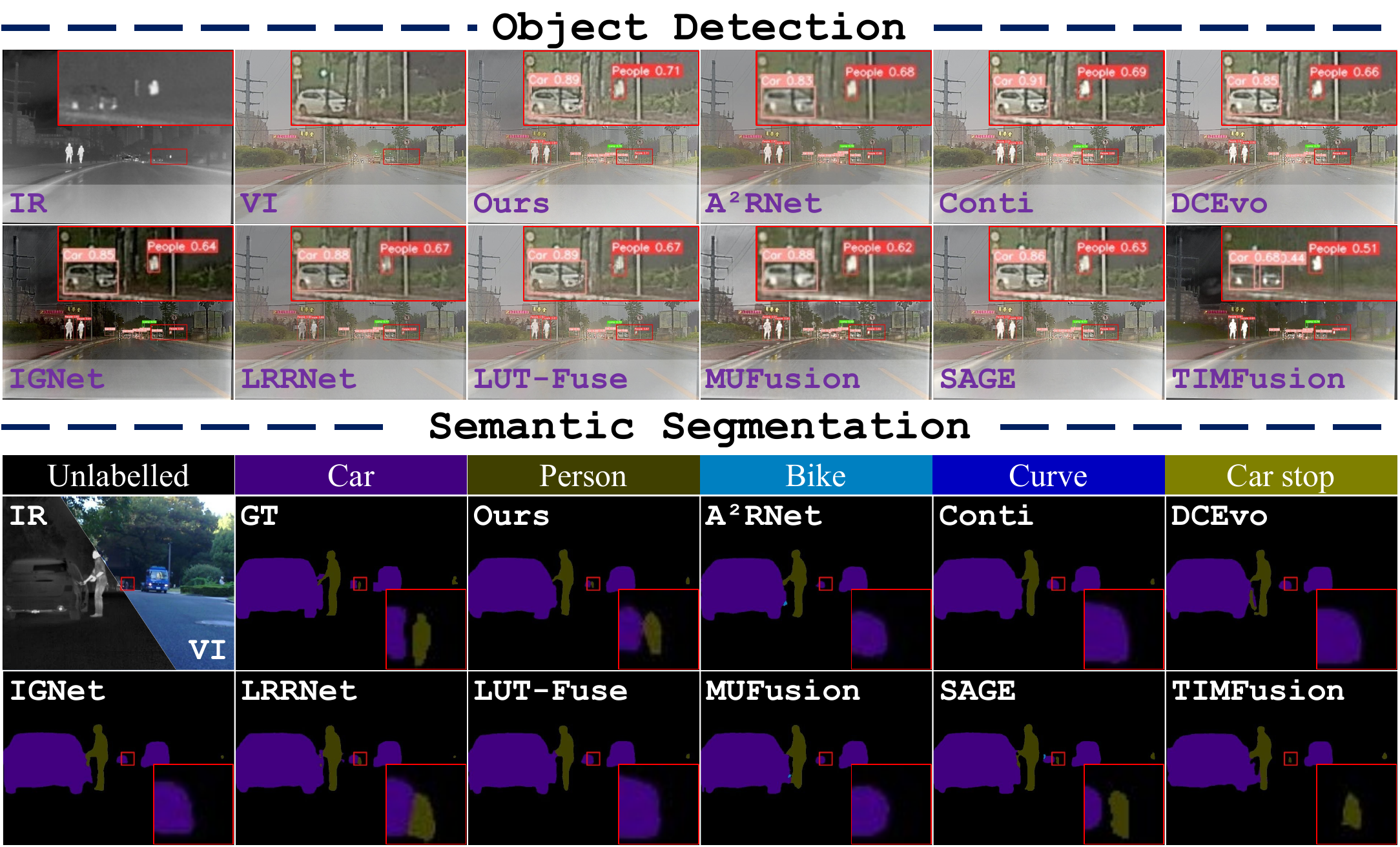}
    \caption{
 Qualitative comparison with SOTA IVIF methods in object detection and semantic segmentation tasks.
    }
    \label{Object}
\end{figure}


\subsection{Performance on High-Level Vision Tasks}
To further validate the generalization capability of our intervention framework, we evaluate its performance on two representative high-level vision tasks: object detection and semantic segmentation.

\subsubsection{Object Detection}
We follow \cite{38} in splitting the M$^3$FD object detection dataset into training, validation, and testing sets with a ratio of 6:2:2. YOLOv5 \cite{48} is adopted as the baseline detector to evaluate detection performance of the proposed method.
As summarized in Table~\ref{tab:dec} and the upper panel of Figure~\ref{Object}, our method attains the highest mAP and the best class-wise AP on \textit{Car} and \textit{Bus}. While methods such as TIMFusion achieve leading AP on specific categories, their substantially lower overall mAP suggests limited cross-category generalization. 
We attribute our gains to CFI's bidirectional cross-modal attention combined with invariance gating, which identifies features that remain informative across different intervention patterns. This yields more complete and robust fused representations. In contrast, correlation-driven methods (\emph{e.g.}, LRRNet, DCEvo) often produce blurred thermal structures or insufficient local contrast, ultimately reducing detection accuracy.

\subsubsection{Semantic Segmentation}
Following \cite{38}, we adopt the MSRS dataset and employ the official segmentation network \cite{37} to perform the semantic segmentation task, thereby evaluating the performance of the proposed method in segmentation. Table \ref{tab:seg} presents quantitative results across six categories. Our method attains the best or near-best performance on multiple key classes, demonstrating balanced generalization across semantic categories. The lower panel of Figure \ref{Object} shows that our method produces finer delineation of vehicles and pedestrians than SOTA methods. 
This advantage stems from our random masking intervention during training: by corrupting identical regions across modalities, the model learns to reconstruct complete semantics from partial evidence, thereby achieving higher detail fidelity in regions with complex boundaries or delicate textures.

\begin{table}[t!]
\centering
\caption{Quantitative results of our proposed method \emph{vs.} SOTA methods on the object detection. The best value is highlighted with {\textbf{Bold}}.}
\label{tab:dec}
\small
\setlength{\tabcolsep}{3pt}
\begin{tabular}{l*{7}{c}}
\toprule
\textbf{Method} & \textbf{mAP} & \textbf{Peo} & \textbf{Car} & \textbf{Bus} & \textbf{Mot} & \textbf{Tru} & \textbf{Lam} \\
\midrule
TIMFusion  & 0.796 & \textbf{0.802} & 0.900 & 0.846 & 0.653 & 0.816 & 0.756 \\
Conti & 0.809 & 0.789 & 0.910 & 0.875 & 0.650 & 0.834 & 0.797 \\
SAGE       & 0.815 & 0.783 & 0.908 & 0.870 & \textbf{0.704} & 0.813 & 0.814 \\
MUFusion   & 0.804 & 0.798 & 0.906 & 0.881 & 0.660 & 0.792 & 0.788 \\
LUT-Fuse   & 0.804 & 0.772 & 0.905 & 0.881 & 0.654 & \textbf{0.836} & 0.775 \\
LRRNet     & 0.811 & 0.780 & 0.911 & 0.878 & 0.694 & 0.804 & 0.798 \\
IGNet      & 0.673 & 0.767 & 0.855 & 0.739 & 0.452 & 0.662 & 0.565 \\
DCEvo      & 0.809 & 0.780 & 0.907 & 0.891 & 0.675 & 0.788 & \textbf{0.815} \\
A$^2$RNet     & 0.809 & 0.795 & 0.906 & 0.886 & 0.659 & 0.810 & 0.798 \\
\midrule
\textbf{Ours} & \textbf{0.821} & 0.800 & \textbf{0.916} & \textbf{0.894} & 0.693 & 0.812 & 0.809 \\
\bottomrule
\end{tabular}
\end{table}

\begin{table}[t!]
\centering
\caption{Quantitative results of our proposed method \emph{vs.} SOTA methods on the semantic segmentation. The best value is highlighted with {\textbf{Bold}}.}
\label{tab:seg}
\small
\setlength{\tabcolsep}{3pt}
\begin{tabular}{l*{7}{c}}
\toprule
\textbf{Method} & \textbf{Unl} & \textbf{Car} & \textbf{Per} & \textbf{Bik} & \textbf{Cur} & \textbf{Roa} & \textbf{mIOU} \\
\midrule
TIMFusion  & 0.975 & 0.826 & 0.597 & 0.597 & 0.389 & 0.450 & 0.639 \\
Conti & 0.982 & 0.878 & 0.699 & 0.671 & 0.540 & 0.646 & 0.736 \\
SAGE       & 0.982 & 0.871 & 0.678 & 0.680 & 0.548 & 0.621 & 0.730 \\
MUFusion   & 0.981 & 0.866 & 0.667 & 0.679 & 0.507 & 0.634 & 0.722 \\
LUT-Fuse   & 0.982 & 0.878 & 0.687 & 0.681 & 0.564 & 0.623 & 0.736 \\
LRRNet     & 0.982 & 0.880 & 0.676 & 0.679 & 0.548 & 0.642 & 0.734 \\
IGNet      & 0.982 & 0.872 & 0.685 & 0.688 & 0.541 & \textbf{0.648} & 0.736 \\
DCEvo      & 0.982 & 0.878 & 0.687 & 0.679 & 0.524 & 0.638 & 0.731 \\
A$^2$RNet     & 0.982 & 0.881 & 0.687 & \textbf{0.688} & 0.556 & 0.644 & 0.740 \\
\midrule
\textbf{Ours} & \textbf{0.983} & \textbf{0.883} & \textbf{0.707} & 0.686 & \textbf{0.584} & 0.642 & \textbf{0.747} \\
\bottomrule
\end{tabular}
\end{table}

\subsection{Medical Image Fusion}
\begin{figure}[t!]
    \centering
    \includegraphics[scale=0.26]{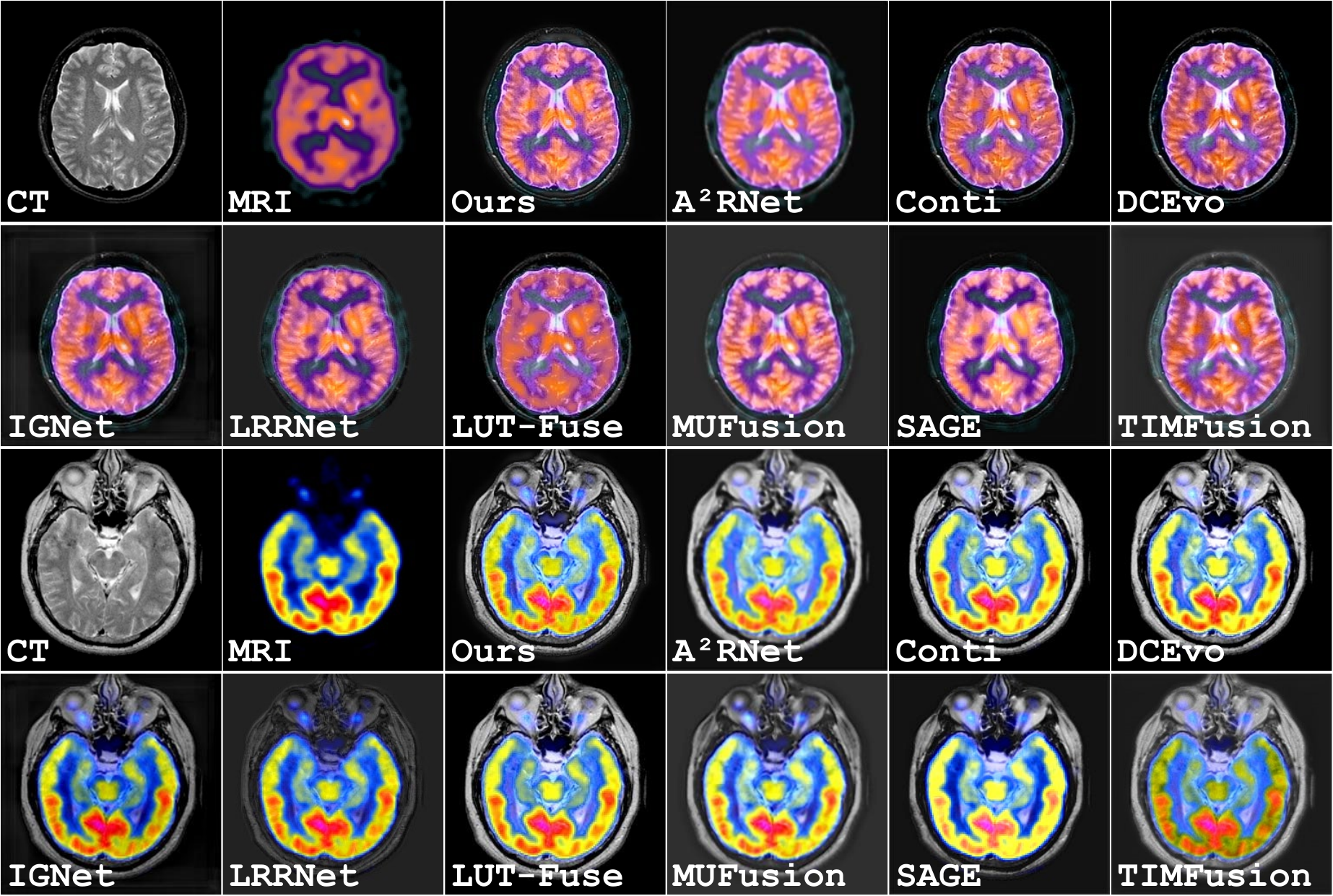}
    \caption{
Qualitative comparison with SOTA methods on the MIF.
    }
    \label{Medical}
\end{figure}

To evaluate cross-domain generalization, we conduct medical image fusion (MIF) experiments on the Harvard medical dataset \cite{47} containing 20 MRI-PET/SPECT image pairs. Notably, we directly deploy the IVIF-trained model to MIF without fine-tuning, constituting a challenging cross-sensor transfer.
Table \ref{tab:medical} shows our method achieves the highest AG and SF with competitive PSNR and CC. Figure \ref{Medical} further demonstrates that our method produces coherent contours and stable contrast for fine anatomical structures, while correlation-driven methods like LRRNet and DCEvo yield blurred details, and semantically-guided methods like TIMFusion and SAGE exhibit over-sharpening artifacts or texture collapse.
This cross-sensor transfer success validates that intervention-based training captures fundamental fusion principles rather than domain-specific patterns. The invariance gating mechanism successfully identifies features that remain informative across vastly different imaging modalities, from thermal-visible to anatomical-functional medical imaging, confirming the generalizability of intervention-stable features.

\begin{table}[t!]
\centering
\caption{Quantitative results of our proposed method \emph{vs.} SOTA methods on the MIF. The best value is highlighted with {\textbf{Bold}}.}
\label{tab:medical}
\small
\setlength{\tabcolsep}{8pt}
\begin{tabular}{l*{5}{c}}
\toprule
\textbf{Method} & \textbf{AG} & \textbf{SF} & \textbf{PSNR} & \textbf{CC} & \textbf{$\mathcal{Q}_{abf}$} \\
\midrule
TIMFusion  & 5.122 & 5.840 & 61.37 & 0.800 & 0.467 \\
Conti & 7.205 & 9.473 & 63.98 & 0.857 & 0.637 \\
SAGE       & 5.406 & 6.860 & 63.91 & 0.861 & 0.376 \\
MUFusion   & 4.777 & 5.234 & 62.21 & 0.861 & 0.355 \\
LUT-Fuse   & 6.549 & 8.467 & 64.34 & 0.852 & 0.605 \\
LRRNet     & 3.753 & 4.564 & 63.92 & 0.834 & 0.211 \\
IGNet      & 5.623 & 6.035 & \textbf{65.13} & 0.860 & 0.499 \\
DCEVO      & 7.391 & 9.585 & 63.82 & 0.850 & \textbf{0.688} \\
A$^2$RNet     & 5.131 & 5.317 & 63.18 & 0.860 & 0.386 \\
\midrule
\textbf{Ours} & \textbf{7.681} & \textbf{10.06} & 64.33 & \textbf{0.862} & 0.599 \\
\bottomrule
\end{tabular}
\end{table}

\section{Conclusion}
We propose an intervention-based MMIF framework that actively probes modal dependencies through three complementary masking strategies, testing cross-modal compensation, local feature robustness, and balanced modal utilization respectively. Our CFI employs learnable invariance gating to prioritize perturbation-robust features, steering the network toward genuine dependencies rather than spurious correlations. Experiments demonstrate SOTA performance, cross-sensor robustness, and zero-shot IVIF-to-medical transferability, confirming that our framework captures fundamental fusion principles.




\section*{Acknowledgments}
This work was supported by National Natural Science Foundation of China under Grants No.62162065 and 61761045; Yunnan Province Visual and Cultural Innovation Team No.202505AS350009; Yunnan Fundamental Research Projects No.202201AT070167; Joint Special Project Research Foundation of Yunnan Province No.202401BF070001023.

{
    \small
    \bibliographystyle{ieeenat_fullname}
    \bibliography{main}
}


\end{document}